\documentclass{article}

\usepackage{microtype}
\usepackage{graphicx}
\usepackage{subfigure}
\usepackage{booktabs} 

\usepackage{hyperref}

\usepackage[accepted]{icml2023}

\usepackage{amsmath}
\usepackage{amssymb}
\usepackage{mathtools}
\usepackage{amsthm}

\usepackage[capitalize,noabbrev]{cleveref}

\theoremstyle{plain}

\theoremstyle{definition}

\theoremstyle{remark}

\usepackage[textsize=tiny]{todonotes}

\icmltitlerunning{SLAM Tracking}

\begin{document}

\twocolumn[
\icmltitle{Lidar based 3D Tracking and State Estimation of Dynamic Objects}



\icmlsetsymbol{equal}{*}

\begin{icmlauthorlist}
\icmlauthor{Patil Shubham Suresh}{CMU}
\icmlauthor{Gautham Narayan Narasimhan}{CMU}
\end{icmlauthorlist}

\icmlaffiliation{CMU}{Carnegie Mellon University, Pittsburgh, USA}

\icmlcorrespondingauthor{Patil Shubham Suresh}{shubhamp@andrew.cmu.edu}
\icmlcorrespondingauthor{Gautham Narasimhan}{gautham@andrew.cmu.edu}

\icmlkeywords{Machine Learning, Computer Vision, Autonomous Driving, Object Detection, ICML}

\vskip 0.3in
]



\printAffiliationsAndNotice{} 

\section{Problem Statement}
\label{sec:problem_statement}

3D point cloud data obtained using Lidar sensor has been very crucial in 3D localization and mapping of static objects within a scene. For Autonomous vehicles, Lidar point cloud is fused with camera to help in detecting objects like cars and pedestrians. However, it hasn’t been used to determine the dynamic states like velocity, yaw, yaw rate, etc of nonego objects. The rich positional data obtained using Lidar can be harnessed to track and estimate the relative states of dynamic objects within the scene by using it in conjunction with visual data. Our approach of using Lidar data to track and estimate state information of oncoming vehicles/pedestrians can provide highly parameterized state of each of them. This information can be later used to accurately model and predict future trajectories.
\subsection{Impact and Novelty}
The novel approach used in our project work are:
\begin{itemize}
    \item State estimation of oncoming vehicles: Earlier research has been based on determining states like position, velocity, orientation , angular velocity, etc of ego-vehicle. Our approach focuses on estimating the states of non-ego vehicles which is crucial for Motion planning and decision-making.
    \item Dynamic Scene Based Localization: Our project will work on dynamic scenes like moving ego (self) and non-ego vehicles. Previous methods were focused on static environments.
    \item Lidar Based Odometry: Lidar odometry can complement vehicular odometry estimated using IMU, GPS, wheel odometry etc.
\end{itemize}
\section{Literature Review}
\label{sec:literature}
Currently the most common approach towards determining the states of oncoming vehicles is to use a simple constant velocity model to predict the dynamic states of nonego objects. Generally these values are under constrained
and rely on high co-variance for doing trajectory prediction. Prediction is currently one of the hardest problem in
autonomous vehicles due to lack of state information like position, velocity, acceleration, yaw, yaw rate to correctly
model oncoming vehicle future trajectory as these cannot be
determined accurately using just visual camera data or constant velocity models. Lidar based Odometry has shown
to be quite robust \cite{1} and has been used for mapping \cite{2}
the ego-vehicle surrounding really well, but these have been
implemented in static environments. We also referred to some of the Bird's eye view surveys \cite{s1, s2, a1, s3, s4}  We will be using Argoverse dataset \cite{5} to get Lidar point cloud, camera, egovehicle odometry , etc along with the ground truth.

\begin{figure}[htb]
  \centering
  \centerline{\includegraphics[width=8.5cm]{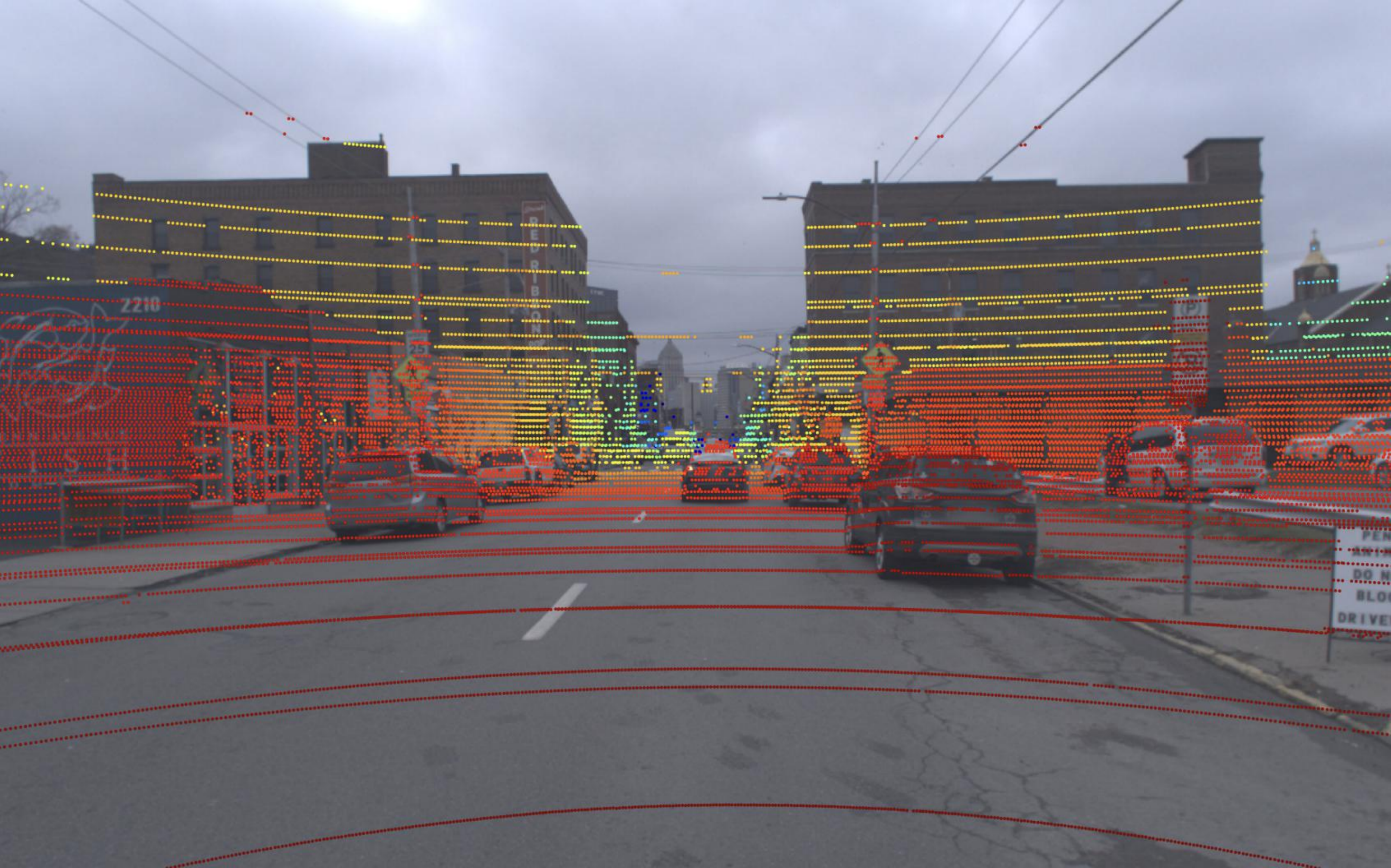}}
\caption{ Lidar points overlapped onto the camera frame}
\label{fig:data1}
\end{figure}

\section{Theory Developed}
In order to achieve the goal of the project, the following
were the main theory along with its implementation.
\subsection{ROS Pipeline}
We have used ROS (Kinetic) framework to communicate
between various modules namely the camera feed and the
corresponding Lidar data.

\subsection{LiDAR-Camera Calibration}
We used extrinsics matrix provided in Argoverse dataset
to get the extrinsic matrix between camera and LiDAR.
Then we projected all the points from pointcloud of LiDAR
on to the camera. We used the inbuilt Argoverse function to
do this. We used intrinsics calibration to convert to images
coordinates. Then we repeated the same step for all the ring
cameras.

\begin{figure}[htb]
  \centering
  \centerline{\includegraphics[width=8.5cm]{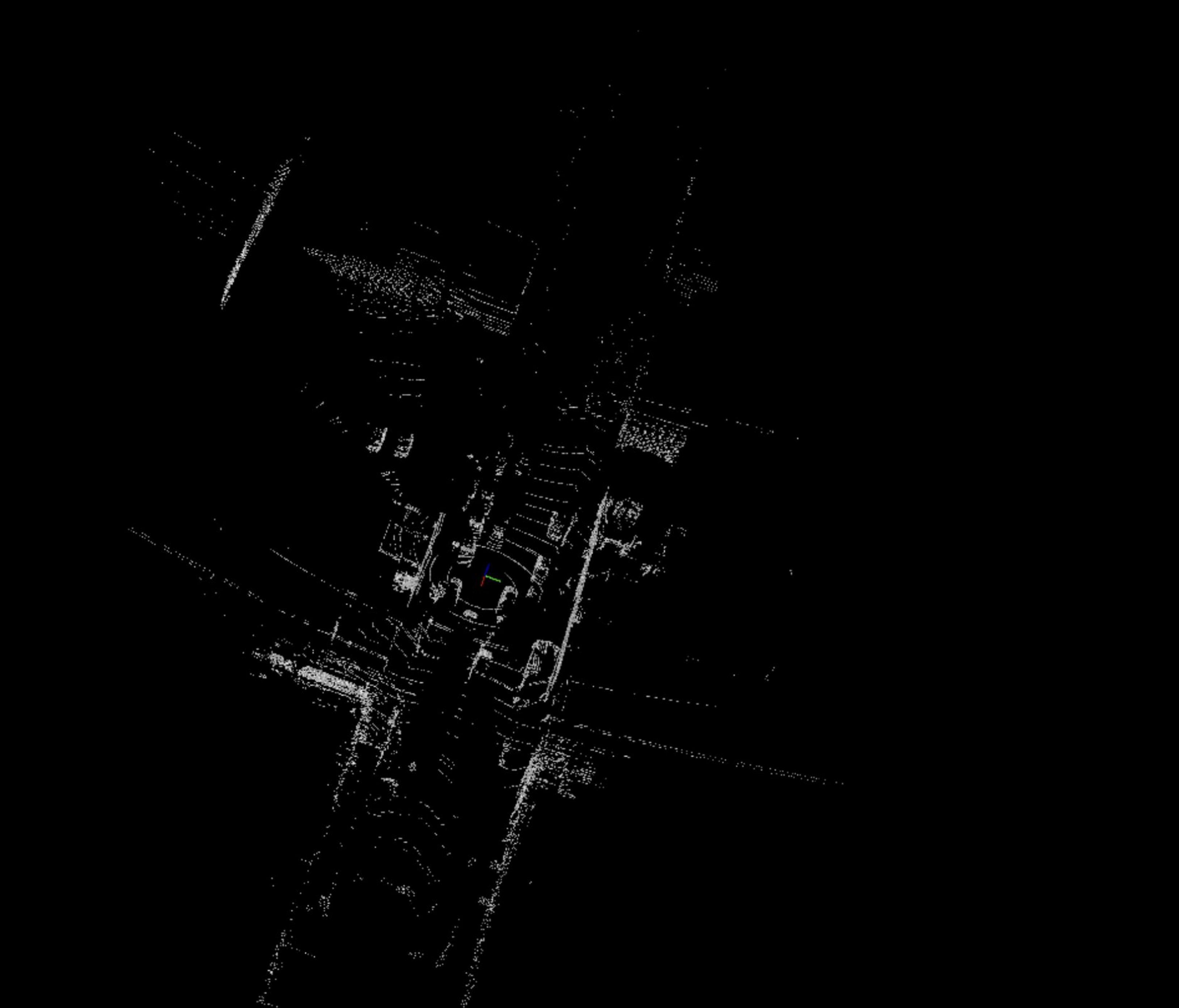}}
\caption{Complete point cloud}
\label{fig:data11}
\end{figure}

\subsection{Noisy points removal using DB-SCAN}
Dense point cloud from LiDAR has as much disadvantages as it has advantages. The first disadvantage being the
computation cost. Number of points in each timestamp is
around 100,000. We need to run clustering on top of this
data, which is O(N*N) time complexity hence it will really
drive the speed of the overall algorithm down. Hence we
need to remove the points. Another disadvantage of dense
point cloud is that we would have lots of noises in addition
to the relevant information since the range of the point cloud
is too large.
Considering all these effects of dense point-cloud we
followed following methods to remove the noise points,
thereby making the point cloud less dense.
\begin{itemize}
    \item Segment out the ground plane: The ground plane is
dense, but does not need to be segmented. We assume
that the Lidar is mounted on the car and the ground
plane is always below the Lidar. We segmented out all
the points below 1.5 meters and then used RANSAC
to fit a plane. The ground plane will be the dominant
plane and all the points lying with the planar RANSAC
are segemented out. The points not lying on the ground
plane are returned back and merged with points above
1.5 meters. Most of these points are those belonging
to the tires of vehicles or feet of pedestrians.
\item Drivable area based points removal: As mentioned
above we only care about the vehicles present on the
road. These vehicles will be present only on the road.
Hence we can visualize the points in the 3D world and
project on drivable area segmentation mask (Provided
with argoverse API). This way we can remove all the
points present outside the drivable area. This will again
help us since the total number of points to be clustered
is reduced.
\item Mask-RCNN based point removal: We didn’t incorporate this method. It was just an experiment to incorporate both the sensor modalities for the point removals.
Here we projected all the 3D-LiDAR points onto the
2D-image. We could easily do this, as shown in the
figure above (through the LiDAR and camera calibration). We the run object detector using Mask-RCNN on the image. Here we had set the confidence threshold for object detection very low to minimize the loss
of important information. Then we only remove all
those LiDAR points which doesn’t lie in any of the
bounding boxes.
\end{itemize}

\begin{figure}[htb]
  \centering
  \centerline{\includegraphics[width=8.5cm]{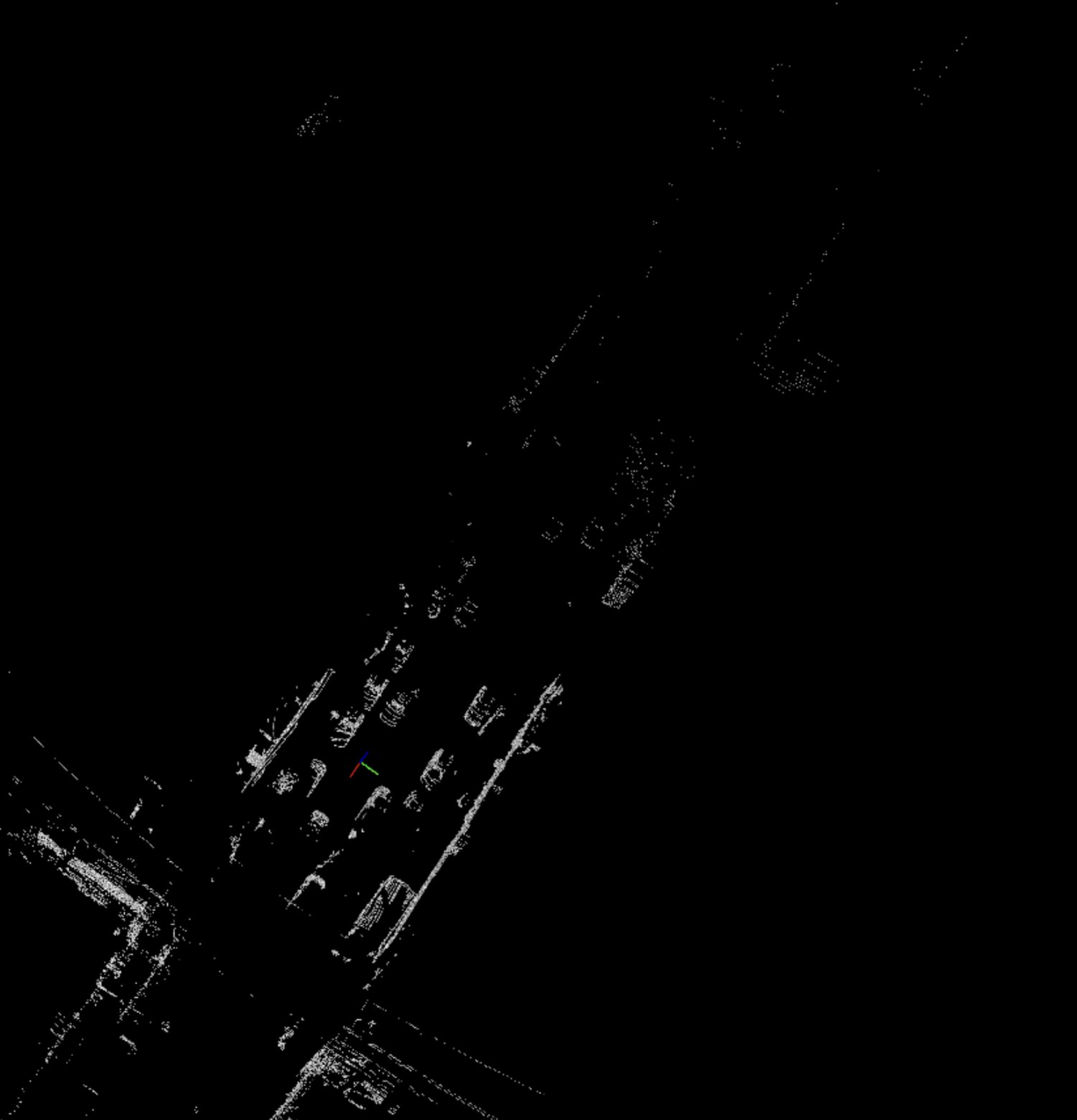}}
\caption{Non drivable area removal}
\label{fig:data10}
\end{figure}

\begin{figure}[htb]
  \centering
  \centerline{\includegraphics[width=8.5cm]{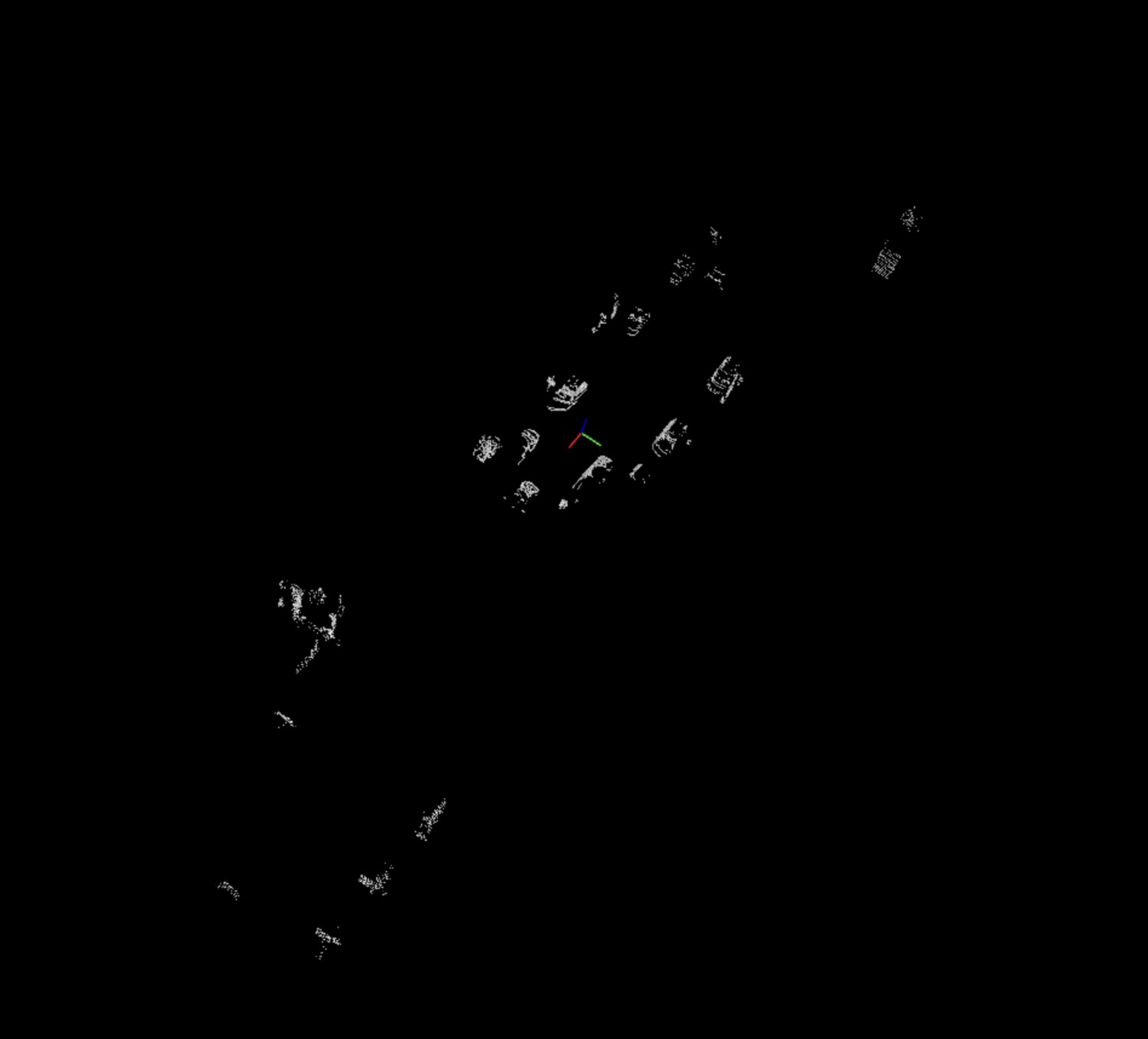}}
\caption{Removal of sparse points}
\label{fig:data9}
\end{figure}

\begin{figure}[htb]
  \centering
  \centerline{\includegraphics[width=8.5cm]{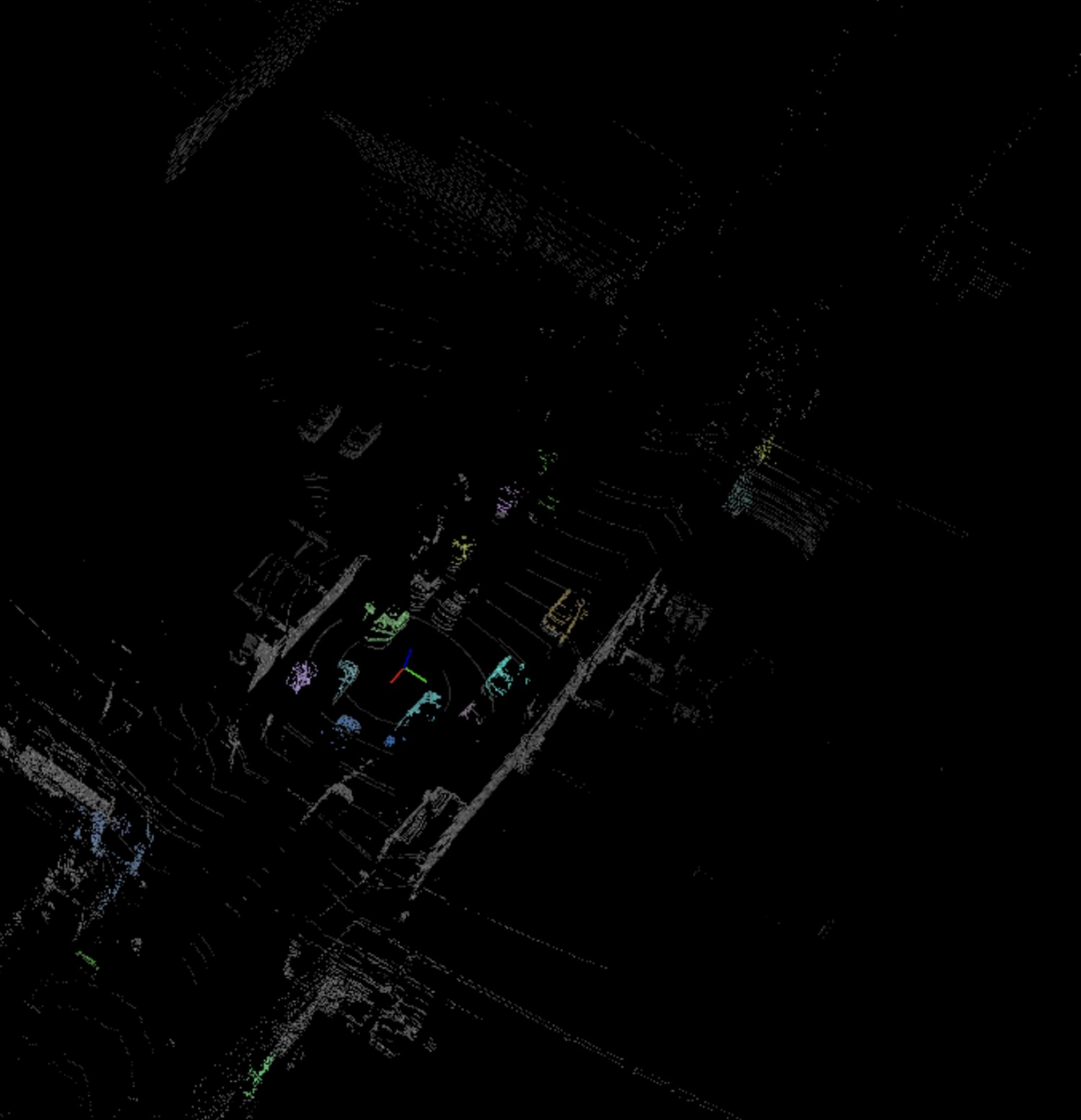}}
\caption{Instance clustering of cars with noises}
\label{fig:data81}
\end{figure}

\begin{figure}[htb]
  \centering
  \centerline{\includegraphics[width=8.5cm]{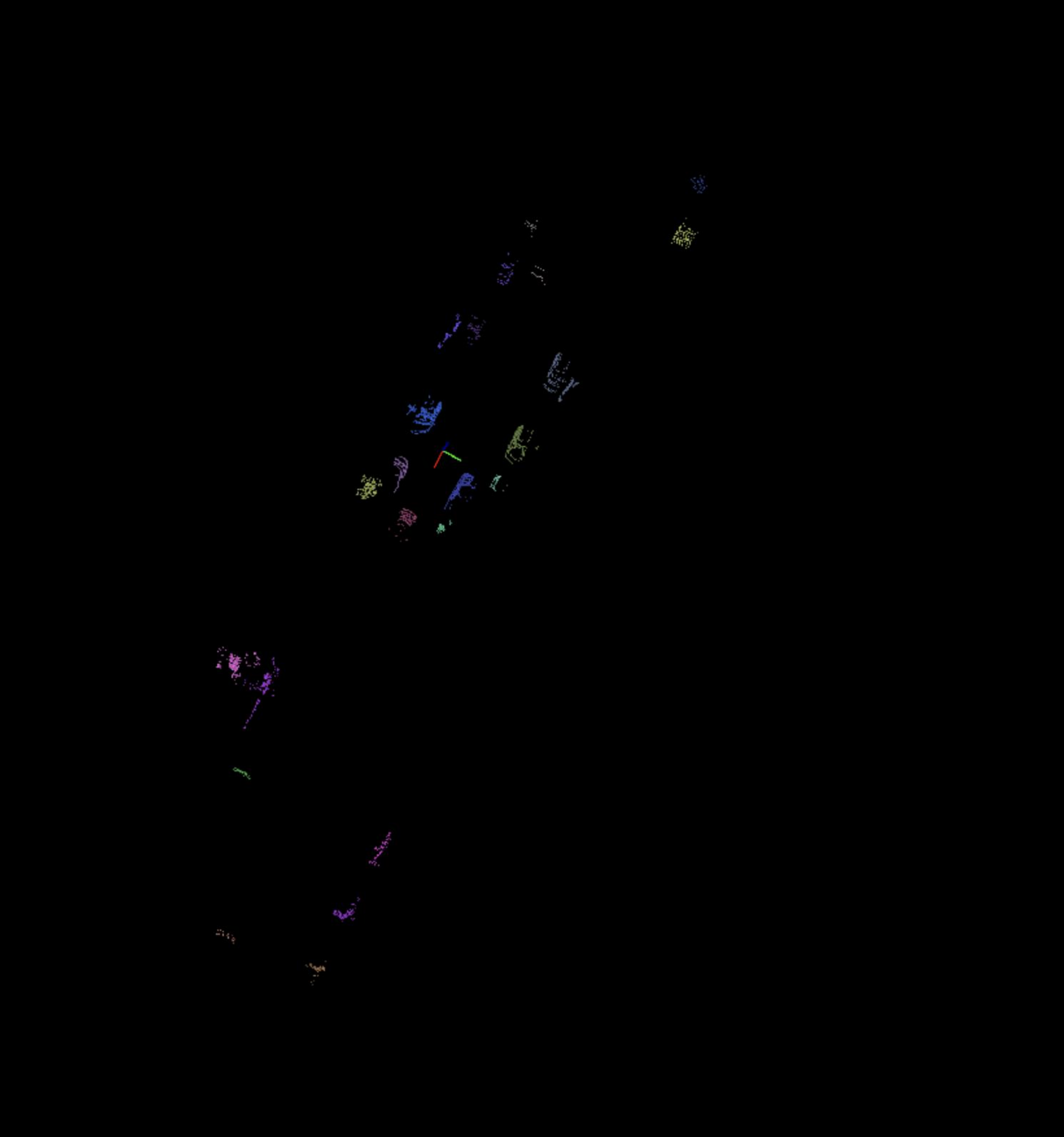}}
\caption{Instance clustering of cars without noise}
\label{fig:data7}
\end{figure}

\subsection{LiDAR Point Cloud Segmentation}
The aim of this task was to design a system that accepts
as input a 3D point cloud and outputs a segmented point
cloud. We followed the following algorithm:
Clustering: We have used Density-based spatial clustering of applications with noise or DB-Scan for clustering.
This clustering algorithm dynamically adapts to the number of points. The algorithm takes in the minimum number
of points to be considered a cluster. This serves our needs
better than K-Means clustering which requires fixed seed
initialization. The was part of our real time requirement for
processing. We intend to have a frame rate of 2. This DBScan clustering that we used also in turn removes the noisy
points - that don’t qualify for being dense enough. Essentially we would just need to pass in two parameters to the
DB-Scan function - Epsilon and min points. Epsilon defines how far do we look for the neighbours and min points
defines that how many points should be there at minimum
with the distance less than epsilon.
\subsection{3D-Bounding Box creation}
3D-Bounding boxes were created around the segmented
point clouds. We used the extreme corner points of a cluster
to fit the bounding box. We are doing this in 3D space, since
objects in the 3D space have consistent size. We further
enforced some heuristics based methods to enforce the size
and shape of these bounding boxes. To elaborate that we put heuristics on 4 things - height, width, length, and area.
We put upper and lower bounds on them. Those heuristics
were based on the actual car dimensions that we generally
see on roads. In addition to this we are working on getting
the centroid of the vehicle for a better visualization.
\subsection{Initializing tracks/Dropping old tracks}
Here, by default we are creating new tracks instances
whenever we are finding an unmatched detections (detections that is not associated to any of the active track instances). Such method will also initiate tracks for ghost detections. Hence we came up with two parameters to penalize the tracks initiated by ghost detections. Firstly, we use
hit threshold. This is a variable of the track instance. We
increment this value whenever the detection is the current
time stamp is matched with the previously initiated track
instances. We send the track instance to the visualization function only if the current hit value is greater than 5 i.e.
we have done detection and data association of a car in at
least 5 frames.
Another threshold that we use is miss threshold. This
threshold is defined to drop the tracks, for which the detection is not available for the long enough time. We have
set this threshold = 5. This means that if the detection is not
associated in continuous 5 frames then we will just delete
that particular instance of the track.
\subsection{Point Cloud Isolation}
Earlier, we had decided on using Frustum-Net to isolate
point clouds that lie at the front of the vehicle. It uses RGBD information. We decided to instead use Mask R-CNN
based approach to detect point clouds of interest lying in
front of the vehicle. We first used Mask R-CNN region proposal network to get the possible object proposals on the
image. Through this we get object mask in pixel space. According to our experience there were around 100-150 such
regions in an image. There were multiple images per frame
(around the vehicle). Then since we have camera and LiDAR already calibrated, we segmented out all those points
from the point-cloud that were not lying inside the regions
proposed from Mask R-CNN. This step provided two functionalities - 1) It decreased the amount of noises present in
the environment (since we only care about oncoming vehicles for our problem). 2) Less points means less computation time for the functions further down the pipeline.
\subsection{Motion compensation and Motion Model}
While doing data association it is very important to compensate for the ego-motion. Here we predict the motion of
the active tracks through Kalman Filter and then predict the
motion of those tracks and do motion compensation (information of which is already present in the Argoverse data-set
- which essentially came from the odometer sensor). Here
to predict the motion of the active tracks we experimented
on multiple motion models like static, constant velocity and
constant acceleration. We then fixed our model on constant
velocity which gave the best result and least complexity.
\subsection{Data Association}
For data association we use off-the-shelf Hungarian algorithm \cite{6}. Its an algorithm to associate the data by minimising the overall cost. This implementation returns the
indices of the two set of detections and also it tells those
detections which were not able to get associated in both the
sets.
\section{Dataset Used}
We are using Argoverse \cite{5} dataset for evaluating our
project. We chose this dataset, because of the availability of the following: odometry information, camera inputs, LiDAR inputs, on-coming vehicle future trajectories. All of
this information is important for testing and evaluation of
our approach. We also get the projection of Lidar point
clouds onto the camera frame. Results shown in figure
6 represents baseline visualization of bounding box using
classical methods (clustering) on LiDAR point-cloud using points removal through various techniques as described
above.
\begin{figure}[htb]
  \centering
  \centerline{\includegraphics[width=8.5cm]{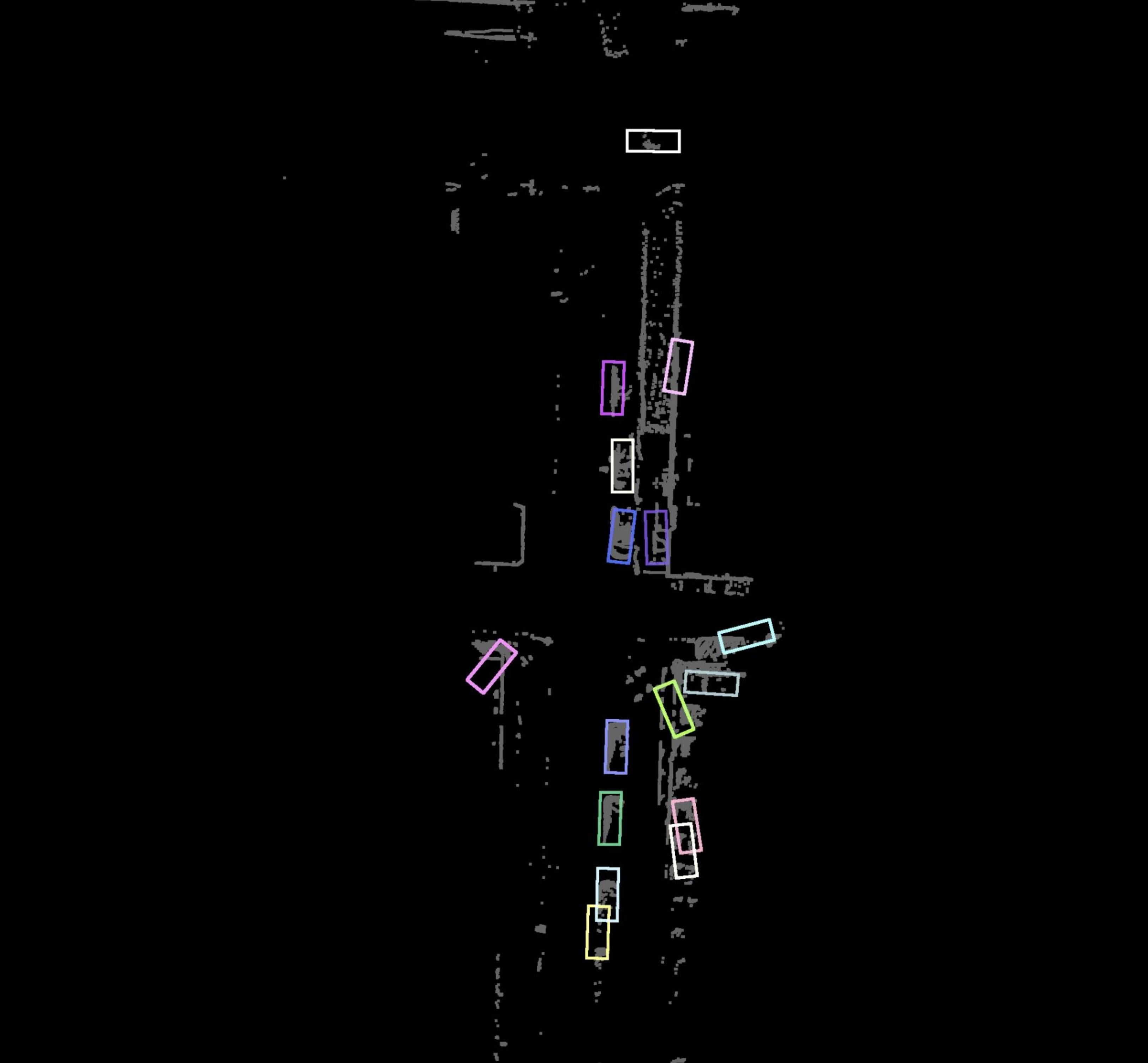}}
\caption{Visualization of 3D bounding box in bird’s eye view using baseline implementation (a)}
\label{fig:data8}
\end{figure}

\begin{figure}[htb]
  \centering
  \centerline{\includegraphics[width=8.5cm]{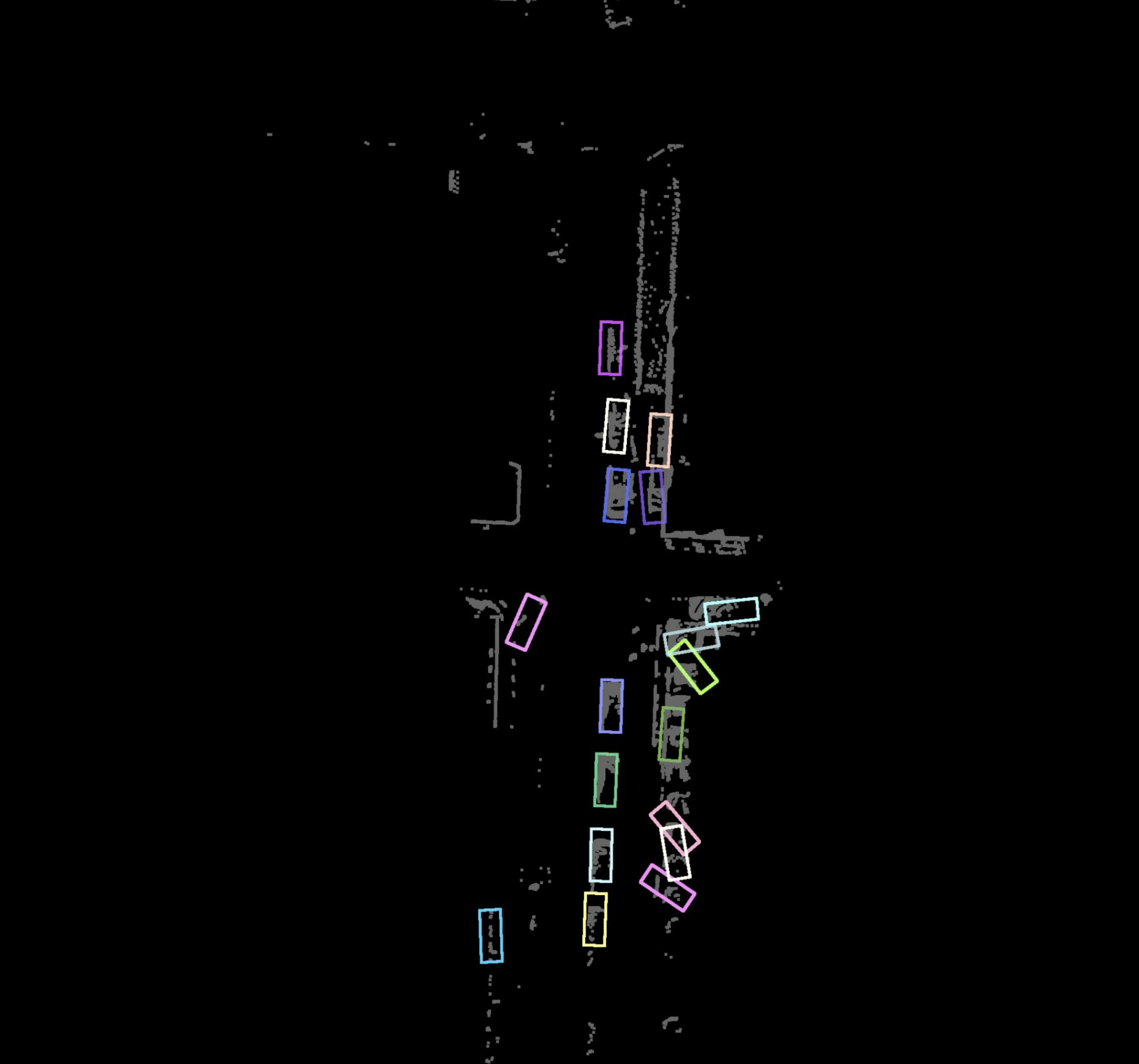}}
\caption{Visualization of 3D bounding box in bird’s eye view using baseline implementation (b)}
\label{fig:data6}
\end{figure}

\begin{figure}[htb]
  \centering
  \centerline{\includegraphics[width=8.5cm]{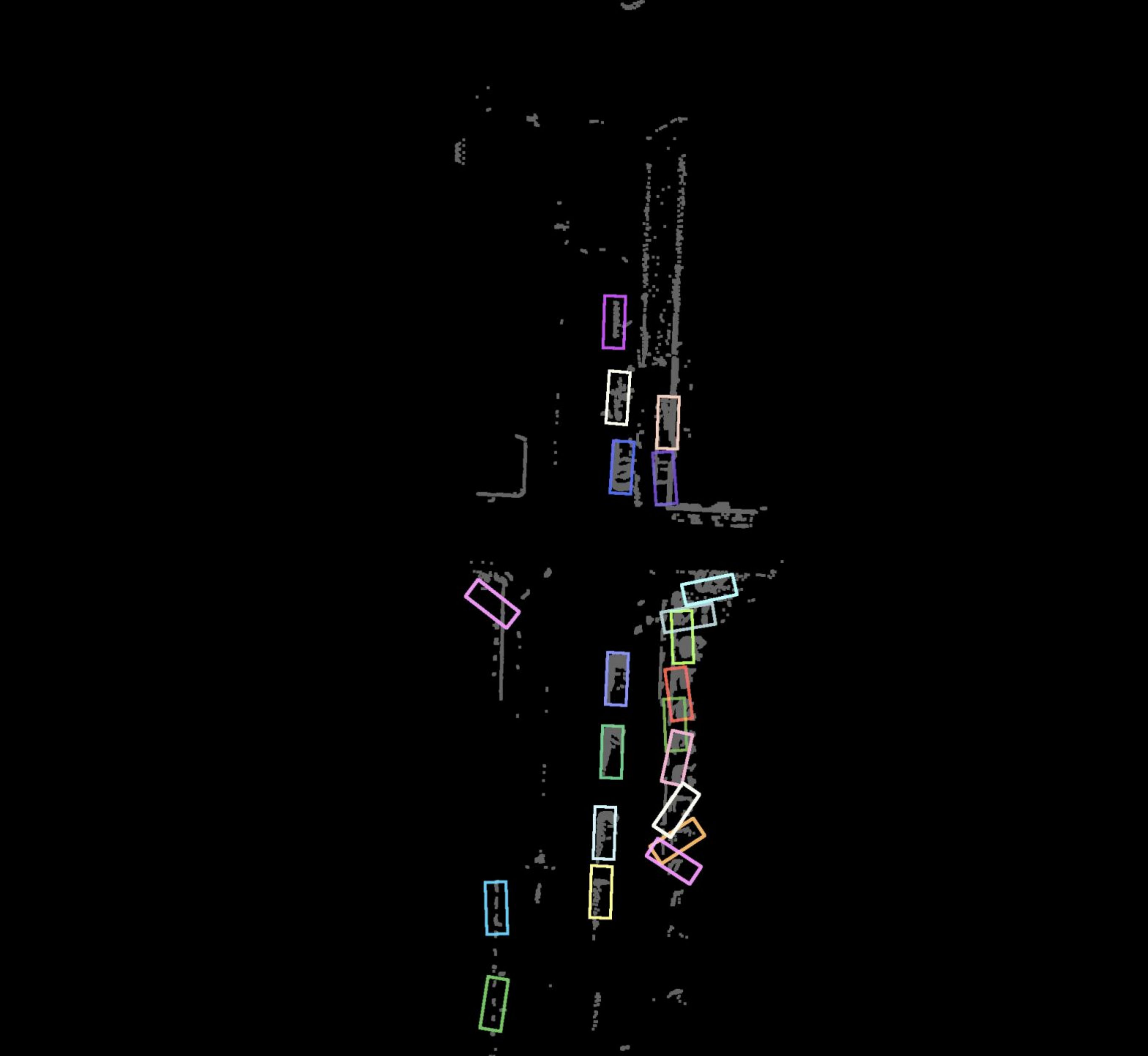}}
\caption{Visualization of 3D bounding box in bird’s eye view using baseline implementation (c)}
\label{fig:data5}
\end{figure}

\section{Code Implementation}
The implentation is done in Python 3.6 and ROS Kinetic.
Initially, we had decided on using C++ due to performance
overheads in Python. However, as didn’t have any real time
validation requirement and Python is great for protoyping,
we decided to stick to it.
For clustering we used Scikit-Learn library with DBScan which is built into it. We tried to implement K-d Tree
from scratch but weren’t able to match the performace of
Scikit-Learn in-built function.
Here we followed the baseline implementation pipeline
from argoverse tracking baseline paper. As in the above
mentioned paper’s implementation we used standard computer vision and maths libraries of Python like scipy, Pykitti,
OpenCV, Matplotlib, glob, pickle, Numpy, pcl (Python
binding), Open3D.
\section{Results}
Multi-Object Training Accuracy (MOTA) \cite{7} is used as
to verify the tracking.

\begin{figure}[htb]
  \centering
  \centerline{\includegraphics[width=8.5cm]{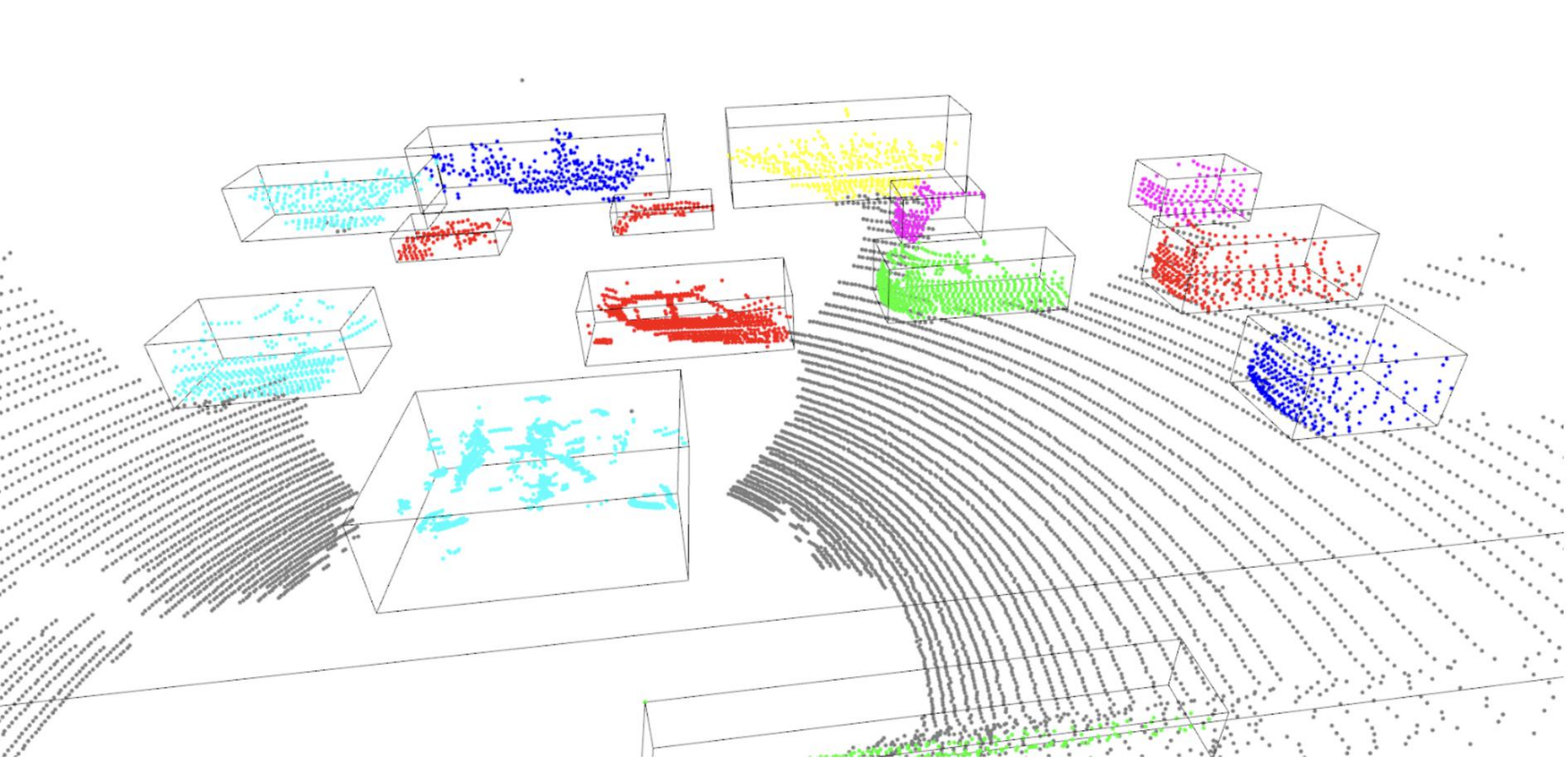}}
\caption{Lidar Point Cloud Segmentation}
\label{fig:data4}
\end{figure}

\begin{figure}[htb]
  \centering
  \centerline{\includegraphics[width=8.5cm]{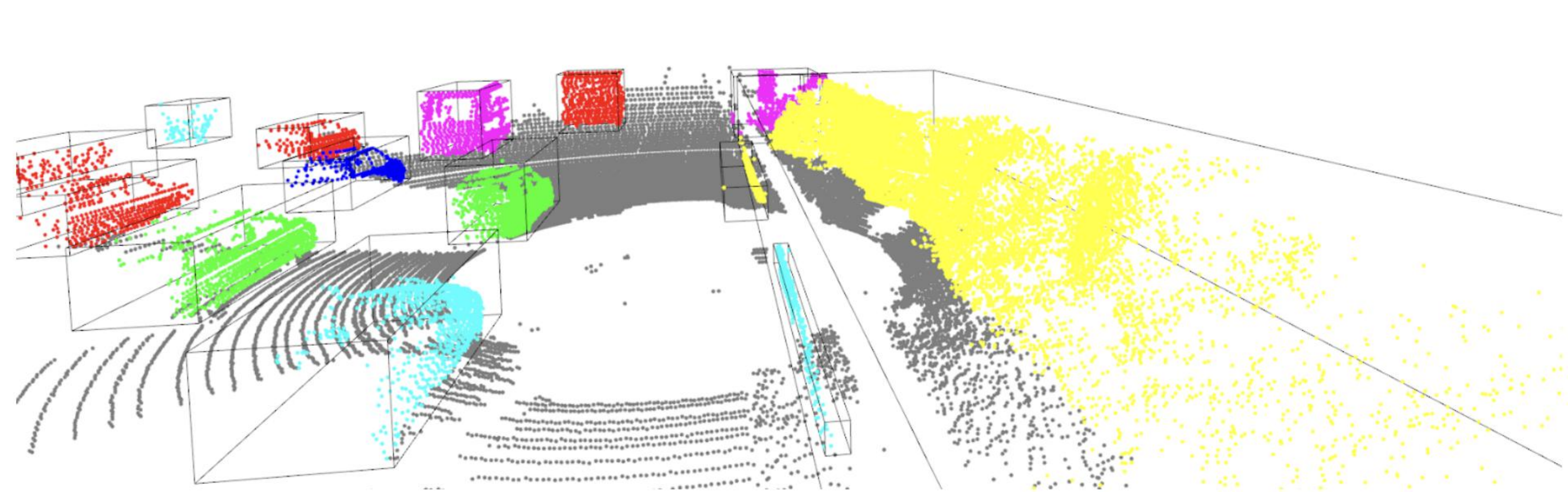}}
\caption{Lidar Point Cloud Segmentation}
\label{fig:data3}
\end{figure}

\begin{figure}[htb]
  \centering
  \centerline{\includegraphics[width=8.5cm]{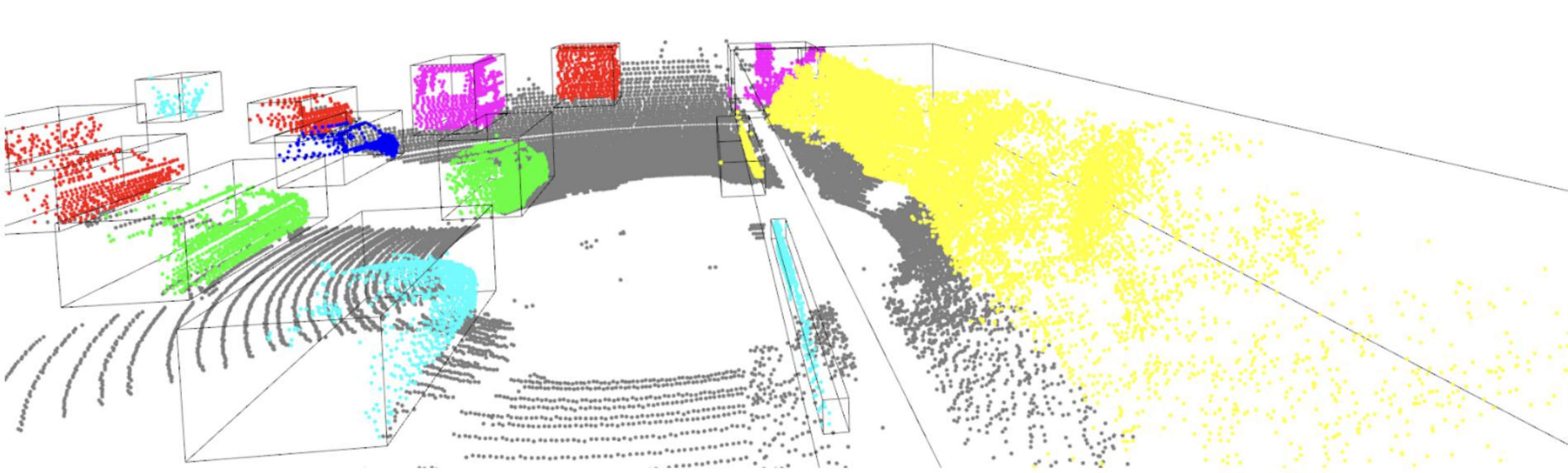}}
\caption{Lidar Point Cloud Segmentation}
\label{fig:data2}
\end{figure}

\section{Challenges faced}
We had planned on using Frustrum Net for isolating
point clouds of targets of interests. It used depth information coming from the camera. The depth information
and will have considerable technical difficulty mapping it
to Lidar points. We instead will be projected points onto
the camera frame and removed points that didn’t lie on the
mask of the object.
We had some challenges in Clustering and segmenting
out the Lidar Point clouds. The point cloud is very dense
(over 10th of a million points per scan). This made computation very expensive. We decided on uniformly reducing
the number of points to 1/10th of the initial value.
Next issue was deciding on a clustering algorithm. KMeans clustering requires seed initialization. As the number of seeds varies with scene, hard-coding the seed value
would not serve the purpose. We then decided on using DBScan which works great for our requirement.
The clustering pipeline was slow (6.02 sec per point
cloud image on a single core). We switched to KD-Tree to
fasten it up. Currently it takes 2.15 per frame with similar
test conditions.
After trying multiple times to run Interative Closest Point
(ICP) to determine the states of on-coming vehicles but no
avail,We believe that Iterative Closest Point matching might
not be able to provide correct homography’s for objects that
are far away and which also have sparser point-clouds describing them. We plan to explore other robust point matching algorithms that may be able to provide an approximate
transform.
\section{Conclusion}
\label{conclusion_section}
We were able to demonstrate that tracking of cars can
be performed using our proposed method and we also
achieved decent results. Specifically, we weren’t able to
implement state estimation due to spare nature of point
cloud and non-rigidity of Iterative Point Methods.We have
lot on our plates to still try. For example we are in process
of replacing parts of our classical methods based function
with deep learning based functions. 

\bibliography{example_paper}
\bibliographystyle{icml2023}



\end{document}